Long Paper*

# A Predictive Model using Machine Learning Algorithm in Identifying Student's Probability on Passing Semestral Course

Anabella C. Doctor
Computer Engineering Department
Lyceum of the Philippines University – Cavite, Philippines
anabellacajolesdoctor@gmail.com
(corresponding author)

## Abstract

*Purpose* – The used of an integrated academic information system in higher education has been proven in improving quality education which results to generates enormous data that can be used to discover new knowledge through data mining concepts, techniques, and machine learning algorithm. This study aims to determine a predictive model to learn students' probability to pass their courses taken at the earliest stage of the semester.

*Method* – To successfully discover a good predictive model with high acceptability, accurate, and precision rate which delivers a useful outcome for decision making in education systems, in improving the processes of conveying knowledge and uplifting student`s academic performance, the proponent applies and strictly followed the CRISP-

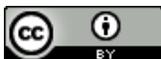

DM (Cross-Industry Standard Process for Data Mining) methodology. This study employs classification for data mining techniques, and decision tree for algorithm.

*Results* – With the utilization of the newly discovered predictive model, the prediction of students' probabilities to pass the current courses they take gives 0.7619 accuracy, 0.8333 precision, 0.8823 recall, and 0.8571 f1 score, which shows that the model used in the prediction is reliable, accurate, and recommendable.

*Conclusion* – Considering the indicators and the results, it can be noted that the prediction model used in this study is highly acceptable. The data mining techniques provides effective and efficient innovative tools in analyzing and predicting student performances. The model used in this study will greatly affect the way educators understand and identify the weakness of their students in the class, the way they improved the effectiveness of their learning processes gearing to their students, bring down academic failure rates, and help institution administrators modify their learning system outcomes.

*Recommendations* – Full automation of prediction results accessible by the students, faculty, and institution administrators for fast management decision making should take place. Further study for the inclusion of some student`s demographic information, vast amount of data within the dataset, automated and manual process of predictive criteria indicators where the students can regulate to which criteria, they must improve more for them to pass their courses taken at the end of the semester as early as midterm period are highly needed.

## INTRODUCTION

In this era, the utilization of technology in education will lead higher institutions to be more efficient in learning their students' at-risk academic performances. The implication of technology in education nowadays leads the educational institutions to become more effective, efficient, accessible, and reliable especially the roles in teaching and student's evaluation performance processes. Hamsa et al. (2016) stated that education is highly significant aspects in regards with the country`s development and higher institutions education quality presented to their students.

Accordingly, one way to accomplish a higher level of quality in educational institutions is the adaptation and implementation of integrated academic information system. An integrated information system nowadays is common in any educational institution, particularly in the integration of academic processes such as test bank, item analysis, and student grading system.  The use of this kind of information system in educational institutions comes up and collected with different kinds of data for both students and faculty such as examination and its results, student information, student class performances, department program curriculum, course syllabus, and the likes.  According to Doctor and Benito (2019), information systems become the pillar of modern education with great significant role contributed to the higher education system nowadays.  It offers countless aid to the educational institutions in managing and storing important high-end information from the students which will become advantageous to the whole education system.

Furthermore, Lay and Nwe (2019) stated that the role of information systems does not only focus to increase the efficacy and efficiency of the organization processes, but also meeting the needs of the students, parents, as well as society.  It is said that integrated campus management systems in higher education are created due to the consideration that it will help administrative processes improved in education particularly in decision making.  The success of managing the education system requires an effective policymaking and system monitoring through the data and information according to Hua et al. (2013).

Due to the adaptation and implementation of an integrated academic information systems in tertiary education, digital format of data has been rapidly growing and highly available. Al-Barrak and Al-Razgan (2016) stated in their study that the needs of data analyzation from the huge data amounts generated form the educational ecosystem will be beneficial to the students and the administrators of higher education.  According to Anuradha and Velmurugan (2015), analyzed data results provides an important factor to the decision-making processes in attaining high quality education.

To produce high quality and relevant data used by school administrators primarily in monitoring students at-risk class performances, a data mining scheme will be needed.  Data mining is the analysis steps of the knowledge discovery in database practice.  It is a way of analyzing data from diverse perspectives and recapitulate it into meaningful and useful information as cited in the study of Nichat and Raut (2017).  Furthermore, data mining which is also known as educational data mining is the process of applying data mining tools and techniques to analyze the data at educational institutions.  Using educational data mining, educational institutions gain thorough and deep knowledge which help them to enhance their assessment evaluation, planning, and decision-making towards their educational curricula. Students' data in educational institutions once analyzed using data mining can give great contributions in academic growth and performance. The main goal of educational institutions is to give the best academic aspects to their learners as well as their overall welfare.  Student tends to undergo a lot of obstacles in life which they need to overcome, for them to achieve optimal academic performance. As the students go on with their education, various types of stress stand as threat towards their learnings and the management office of the institutions handles the situations well enough to help the students cope and get back their academic focus. In the settings of higher education, identifying which students suffered from various obstacles which cause a negative effect to their academic performances is difficult to tell. One key indicator of the students' academic progress is their grades. Making these grades methodically accessible to the students and guidance counselors, efficient method of monitoring students' academic performances will take place. Identifying early learning problems, failure cost and its impacts to the student`s academic performance and knowledge is lesser.

The main goal of this study is to develop an appropriate predictive model using decision tree algorithm to predict the probabilities of the students to pass their courses taken and currently enrolled at the time frame of prelim and midterm period using data mining techniques. This model will be helpful to any higher educations' faculty and administrators to improve the way of conveying knowledge and improve their student's academic performance. The early prediction of students' probability to pass the course by taking analysis academic performances per course can be used by concerned faculty to be alert in helping and guiding their students, and for students, this will make them aware of their current course status. Also, the early detection of the student`s status makes teachers to have more time in preparing interventions appropriate to the learners needs whom needed to be addressed and improve. The outcome of this study can be used in analyzing the effectiveness to the new methods formulated in imparting learnings to the learners, and evaluation methods to help academicians to formulate various models in improving educational policies, guidelines that is effective based on the needs of students.

## LITERATURE REVIEW

### *Education Technology*

Evidently, the role of education takes absolute significant in information society. Büyükbaykal (2015) stated that the future of the country relies on the education of each community the country has, thus the education systems of the country must be restructured by taking account into technological developments. Kurt (2017) stated that "educational technology is a systematic, iterative process of designing instruction or training used to improve performance". It is concerned with the utilization of technology to enhance the way teaching and learning processes in all areas. Joseph (2012) identified in his study that educational technology is concerned about teaching and learning with the involvement of a narrowed range of technology which deal primarily in information and communication that centered on the improvement of teaching and learning process in education.

In the study of Watthananon et al. (2014), it is concluded that the adaptation on the usage of technology in a way of products/applications/tools is to enhance and uplift the way of learning, teaching practices, and instructions with a pure intention of using it as a tool to aid in the delivery of education instead of replacing the current learning practices of an educational institutions have up to these era.

In the current society of learning, learner-centered settings have been the main focused of education reforms. In the study of Joseph (2012), it was strongly stated that when the technology was used to collaborate the traditional teaching and learning, professional development for teacher and student was attained in up high, they become a solution driven oriented, formulate new things, carefully looking back their crafts, handled the information highly, and had a stronger communication and collaboration practices. The implications of using technology in higher education is a means of having a positive and meaningful teaching and learning outcomes towards the learners. Kumar and Vijayalakshmi (2011) cited in the study that technology used to enhance the quality of teaching, to increase access of the learners, and to highly improve the cost-effectiveness of universities and colleges.

*Integrated Academic Information System*

The process of combining one information system of an academic settings, business areas, and others into one is known as integrated system. Integrated system is a system that has been bundled all together to work and function as a single system aiming to efficiently improve productivity and quality of operations around higher education. In universities, the entire academic processes have been transformed into digital through integrated academic information systems utilization. Integrated academic information system is a system composed of different individual system combined and work as one which effectively and efficiently help deliver academic institutions objectives particularly towards their learners. El-Seoud et al. (2017) mentioned that demand for the improvement of integrated systems and integrated digital campus has been continuously and rapidly increasing all throughout and that is why educational technology professionals continue to think more critically about how their students, academics, processes, and technology work together homogenously.

With the transformation and utilization of an integrated systems in many institutions, tons of data have been accumulated and those data becomes stagnant and unused for many years. Additionally, higher institutions before failed to acknowledge the significance of those stored data and how those data would be transformed into a meaningful knowledge towards management decision making. Moreover, as stated in the study of Siddique et al. (2021), the rapid growth of the exploration of data mining towards educational data in various academic purposes has been arising.

*Educational Data Mining*

Data mining is the process of learning a significant patterns from a huge amount of data and identified as emerging and arising structures in providing wide and various of techniques, methods, and tools to enable detailed analysis in different fields. Accordingly, Ahmed et al. (2018) cited that educational data mining (EDM) is also known as data mining in education. Educational Data Mining as describes in a research field with the application of data mining, machine learning and statistics to the information generated from educational settings. Considering the potential application of data mining in educational sector, educational data mining was started as a new stream in the data mining research field.

Furthermore, the interests of educational data mining falls with the new approaches and techniques of inquiring unconventional type of data from educational settings to learn and explore students learning ability. Data mining techniques in educational domains are very useful in improving the current institutions management and educational standards as cited by Anuradha and Velmurugan (2015). Shahiri et al. (2015) found in their study that data mining techniques growing amicably applied in education for developing models to predict students` conducts and performance as e-learning materials in education continuously grows.

Powered by the need of extracting useful information from the large datasets available in educational settings, and discovering hidden patterns and trends, data mining is also known as Knowledge Discovery in Databases (KDD). KDD utilizes variety of statistical and machine learning techniques to analyze data and facilitate decision making. The most common data mining techniques used by researchers to examine the data collected include classification, clustering, association rule mining, regression, and discovery with models. However, according to Papamitsiou and Economides (2014), the main objectives of data mining in educational contexts are student learning behavior modeling, performance prediction, dropout and retention prediction, and identification of students who may be struggling. According to Chrysafiadi and Virvou's study (2013), student modeling is a crucial concept that refers to a qualitative depiction of students' behavior in educational data mining, which can then be utilized to influence instructional decisions.

In a blended learning setting, a predictive model is first fed with raw data from carefully chosen online student interactions before being shifted to useful human activities. Based on the research of Kolo et al. (2015), several different metrics, including IP addresses, content page views, number of quizzes completed, time on task, average session length, messages exchanged/viewed/commented, and content production contributions, can be used to examine the effects of communication, content development, collaboration, and self-evaluation in the effectiveness of a blended learning course.

According to Ebenezer et al. (2019), the data mining process workflow includes numerous phases such as data collection, data preprocessing, and data mining processing. The process of acquiring a significant amount of information is known as data collection, it is essential since a wise choice of qualities will help solve the concept problem. To guarantee that the results of data mining are meaningful, certain data may need to be eliminated or new computed characteristics may need to be added once mining objectives and needs have been established. This feature extraction frequently necessitates the transformation of data recorded in multiple or complex database formats into formats more suitable for the data analysis mining algorithms.

The data cleaning step of data preprocessing also includes error checking and correction, proper handling of null or missing values, and integration of data gathered from various sources into a unified multidimensional format for processing. Finally, as stated by Kaur and Verma (2017), during the data mining processing stage, the features from the structured data set are evaluated using mathematical algorithms that were chosen depending on the type of data and the relationships and patterns the model intends to capture.

The main utilization of data mining, according to Siddique et al. (2021), is to foresee students' academic success. Student academic performance data mining analysis and interpretation are viewed as appropriate analysis, evaluation, and assessment techniques. Certain characteristics, such as academic, personal, family, social, and school characteristics, will be used to predict students' academic performance. According to Siddique et al. (2021), it is determined that early performance prediction would be advantageous for at-risk learners who are having difficulty getting good marks in the class, and that this procedure of predicting really aids such learners in their education to enhance their advancement.

In predicting student academic performance in connection with board examination, there are different algorithms used by the researchers which depend on the nature of their study. In the research by Rustia et al. (2018), classification algorithms (Neural Network, Support Vector Machine, C4.5 Decision Tree, Naive Bayes, and Logistic Regression) were utilized to construct a model by using data mining techniques in predicting the probability of a student passing the Licensure Examination for Teachers (LET). Through machine learning, their study successfully identified which among the classification algorithms used is the appropriate algorithm for this kind of prediction. The researchers used base classifier, multilayer perceptron classifier, J48 classifier, and PART classifier in the research of Siddique et al. (2021). Their investigation concluded that an early performance prediction would support at-risk students who are struggling to achieve good grades in the class. For making predictions, a stable fusion-based ensemble model was created considering students' demographic, family, social, and academic qualities. When evaluating students in the first stages, the model is quite helpful. The best model to forecast students' performance at the lower secondary level appears to be the MultiBoostAB ensemble classifier with MLP basis after creating numerous models involving single, ensemble, and fusion-based ensemble classifiers. Hussain et al. (2019) conducted another study using different algorithms, such as Artificial Immune Recognition System v2.0 and AdaBoost, to predict student performance. This study found that using machine learning techniques to create a predictive model to forecast student performance helped identify underachievers and allowed tutors to take corrective action earlier, even at the start of an academic year, using only internal assessment data from prior semesters, to give at-risk groups more assistance.

### *Machine Learning Algorithm*

To discover significant patterns and models for prediction, data mining employs a variety of algorithms. Data relationships can be categorized into four categories: classes, clusters, associations, and sequential patterns. The performance that may be attained in a certain application domain, the degree of results accuracy, or the model's understandability are some examples of the elements that influence the algorithm to be utilized. According to Ebenezer et al. (2019), classification and clustering may be the mining tasks used the most commonly while mining huge datasets. Because classification is a supervised learning process, the algorithm builds a model that places new, unforeseen cases in established classes after learning from examples that are presented. Each example includes prior knowledge in the form of a value or label designating the class to which the example belongs as well as an input vector of data attributes or characteristics. The base classifiers used to categorize new, unused data by Ebenezer et al. (2019), include nearest neighbors, decision trees, rule-based classifiers, artificial neural networks, support vector machines, and the naive bayes algorithm. Contrarily, clustering is an unsupervised learning process that places data records in groups that already have components that are more like the incoming record than the components in the other clusters. The degree of similarity between items is determined by distance functions, such as Euclidean or cosine distance, and affects the clustering's quality; high-quality clusters have high intra-cluster and low inter-cluster similarity. The three primary clustering techniques utilized in research are hierarchical clustering, K-means clustering, and density-based clustering (Feng, 2019).

Four predictor variables, such as the general weighted average of the students' scores on the mock board, their major, general education, and professional education exams, were defined in the study by Rustia et al. (2018). The respondents were then divided into groups by the researchers, who will now use these groups as study markers. Only the following courses will now be used as predictive variables using the optimum feature selection: English, Science, Theories and Concepts, Methods and Strategies, Special topics, and Core courses. This is the information that will be extracted from the dataset to create the predictive model. According to the findings, out of the five classifiers to be included in the classification model to predict the student's performance on the Licensure Examination for Teachers, only the C4.5 Decision Tree algorithm is taken into consideration.

Multiple Regression models generated the highest accurate prediction results, according to another study of prediction models used in licensure examination performance using classifiers. According to Tarun's (2017) research, the reliability of the Multiple Regression models was demonstrated during testing and simulation using actual institutional data and provided the same results using two trustworthy apps, such as MS Excel and WEKA.

According to the research by Abaya et al. (2016), institutions can anticipate the number of potential board passers from a set of test takers and can also identify the percentage of passing rate with the new prediction method utilizing regression algorithms. In addition, they said that the study's findings showed that variables including age, gender, GPA, and pre-board scores could be used to forecast how well students would perform on license exams.

*Student Class Performance*

Grading systems in an educational institution are very important whose outcomes are coming from various tasks done and performed by the students. The processes of collating and integrating all those things as a result performed by the students inside the classroom is a task which must be done by every teacher of the institutions on a regular basis.

In-depth study is being done in the field of educational data mining, according to Guleria et al. (2014), to forecast student performance to act and prevent failure or dropout. According to Krishna et al. (2020), Grade Point Average (GPA) or grades across assignments, class quizzes and exams, lab work, and attendance, as well as students' demographics, such as gender, age, and family background, and students' individual behaviors, such as beliefs, motivations, and learning strategies, are elements that have frequently been used by researchers in predicting student performance.

According to Shanthini et al. (2018), predicting student achievement is the most efficient way for teachers and students to improve their learning and teaching methods. The classification allows for better inferences, which improves the ability to predict whether a student will pass or fail the courses they are taking. They stated that a model created in prediction will aid the educational institutions to monitor their student's performances in an effective and systematic manner. In addition, they also cited that through the application of data mining algorithms and techniques, the discovered pattern in predicting students' performance will help the educators to understand their learners, identify and discover the weak points of the learners, improve learning processes performed and bring down academic failure rates of the learners.

Predicting student performance gives helpful insights to the teachers particularly to the students who performed less in the class, on what necessary actions and assistance they can provide in terms of learning. The results of prediction provide helps to the institutions to change the and adjust the factors that contributed towards students past poor performance inside the classroom as stated by Krishna et al. (2020).

## METHODOLOGY

Design methodology was used to achieve effective and desirable results in any research. This study employed the Cross Industry Standard Process - Data Mining (CRISP-DM), which is made up of various stages that the researcher must use and go through to achieve the study's goal. This methodology offers a well-structured data mining approach for planning a data mining project in the analytical processes and perspective cited in Gatpandan and Ambat's study (2017).

### *Business and Data Understanding*

The main objective of this study is to come up / provides a predictive model that will predict students' probabilities on passing the courses they currently taken within the current semester in the time frame of prelim up to midterm period. The predictor variables and predicted variable used in this study has been extracted in the database of an integrated academic information system.

To understand and view the overall concepts where the predictions and predicted variables used in this study, the researcher provides the system architecture and data flow diagrams of each academic main modules involved in the academic integrated system used in this study such as examination management system, examination item analysis, and student grading system and three other sub-main modules namely: faculty loading, curriculum management, and syllabus management system.

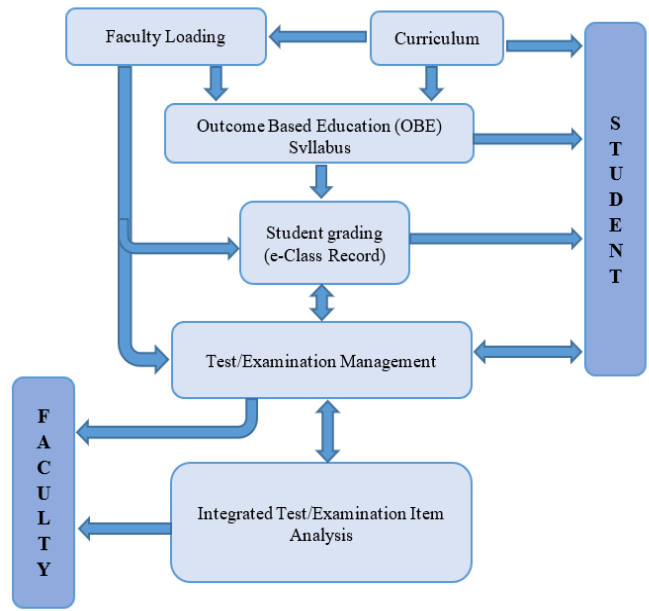

*Figure 1: System Architecture of the Integrated Academic Information Systems*

this study. It gives the overall concepts of the different components or modules the integrated systems comprised and how they interrelated with each other.

To show the detailed picture on how each component comprised in this integrated system interact with each other, Figure 2a-2e take place. The given diagrams show the detailed visual analysis of the whole data of integrated academic systems used in this study. These diagrams present the flow of data and show how information input to and output from the system, the resources, and destinations of that information, and where that information stored. In addition, these figures will give the whole picture to the reader of this study to visualize the fields/variables used in the dataset that has been mined by the researcher to obtain the goal of predicting students' probabilities on passing the courses they currently have taken.

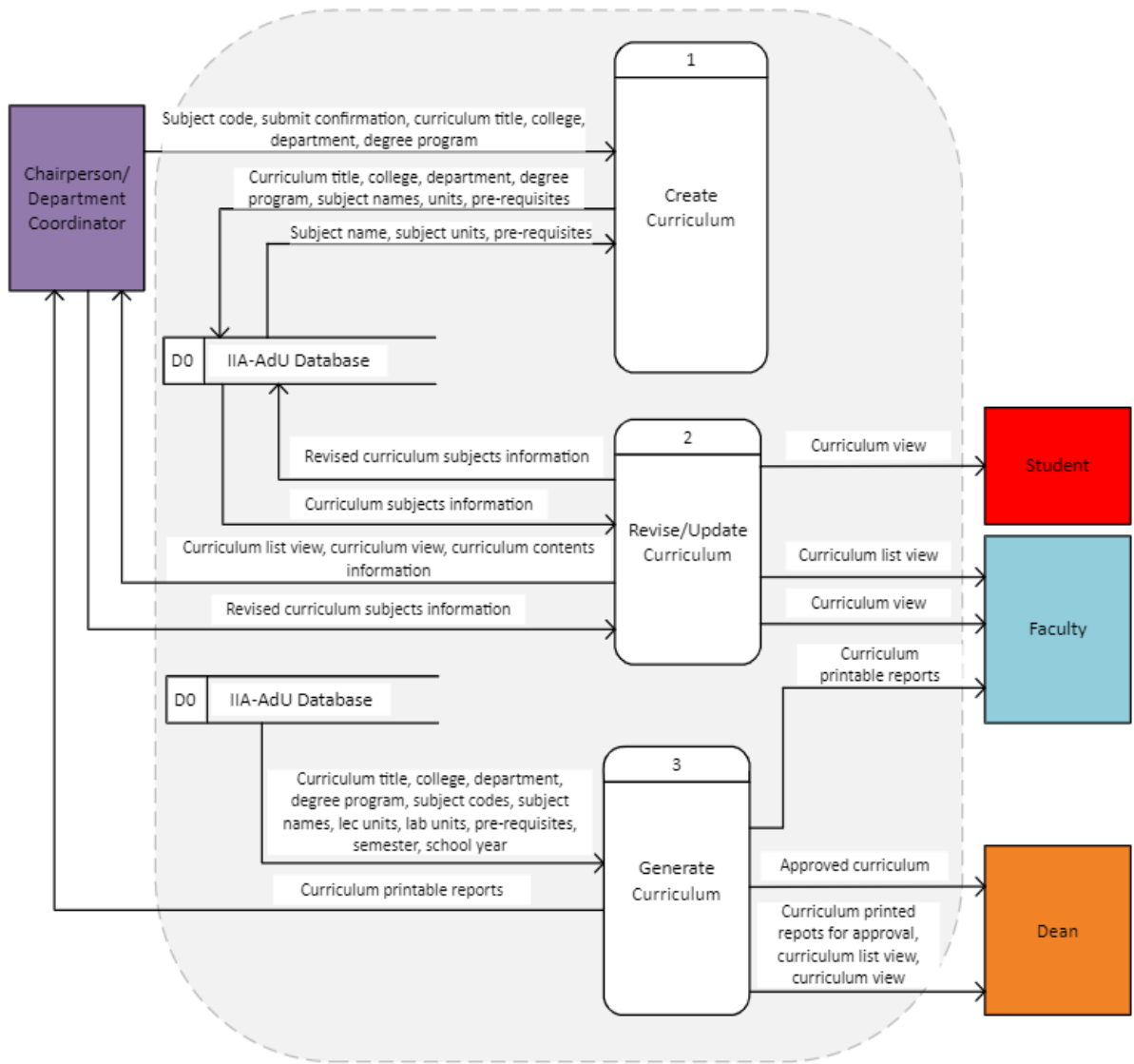

*Figure 2a.* Data Flow Diagram of Curriculum Management Module

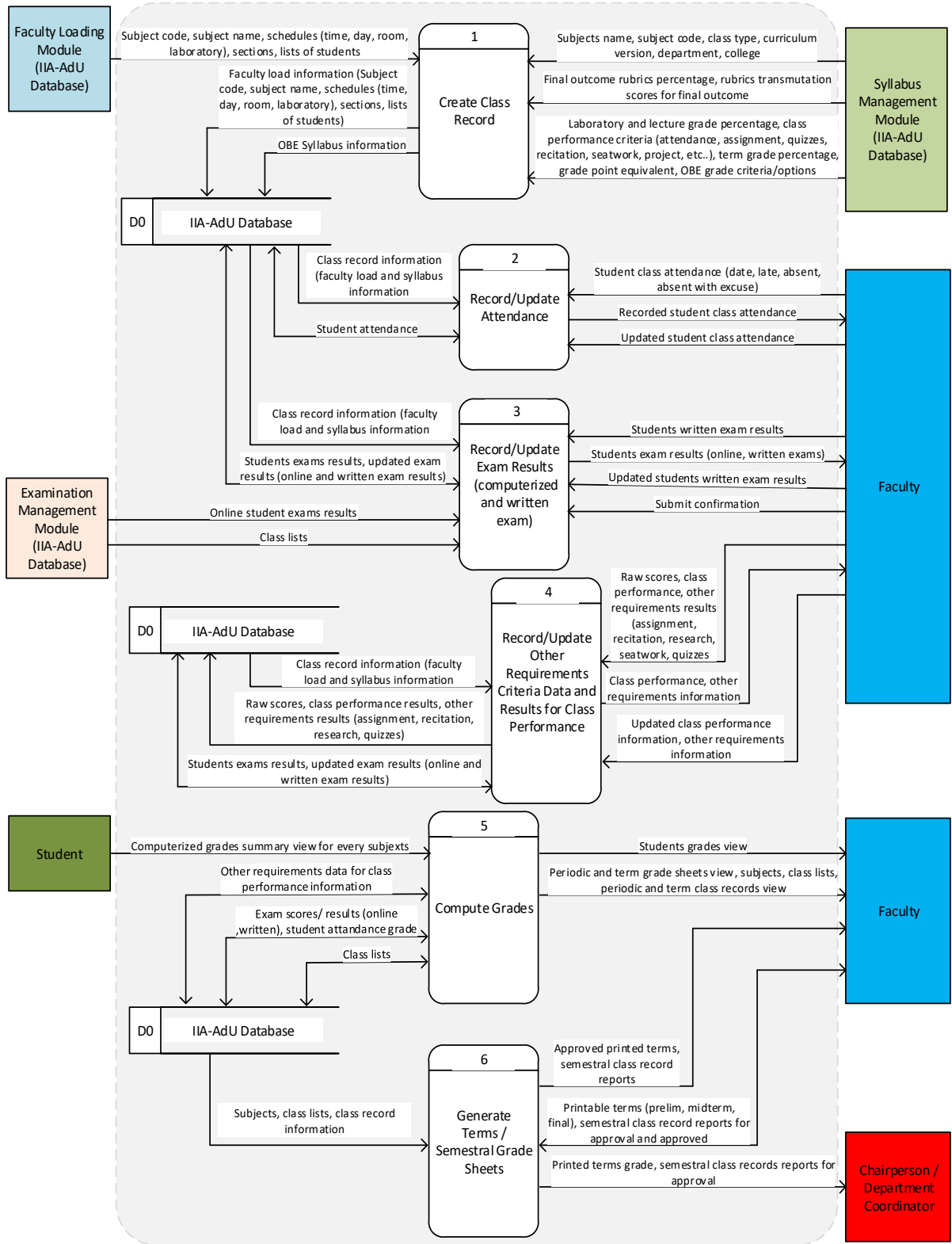

*Figure 2b.* Data Flow Diagram of Student Grading System Management Module

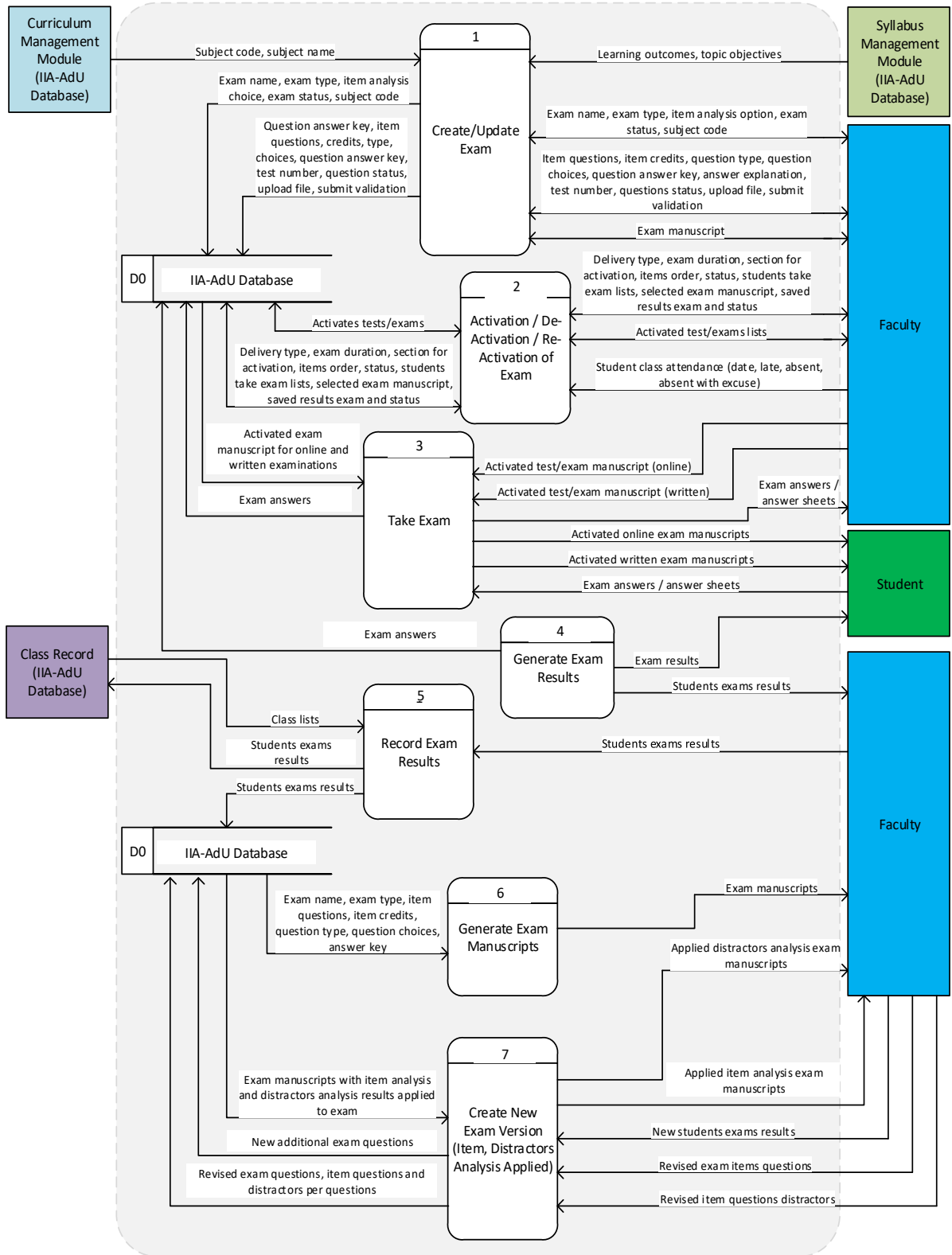

*Figure 2c.* Data Flow Diagram of Examination Management Module

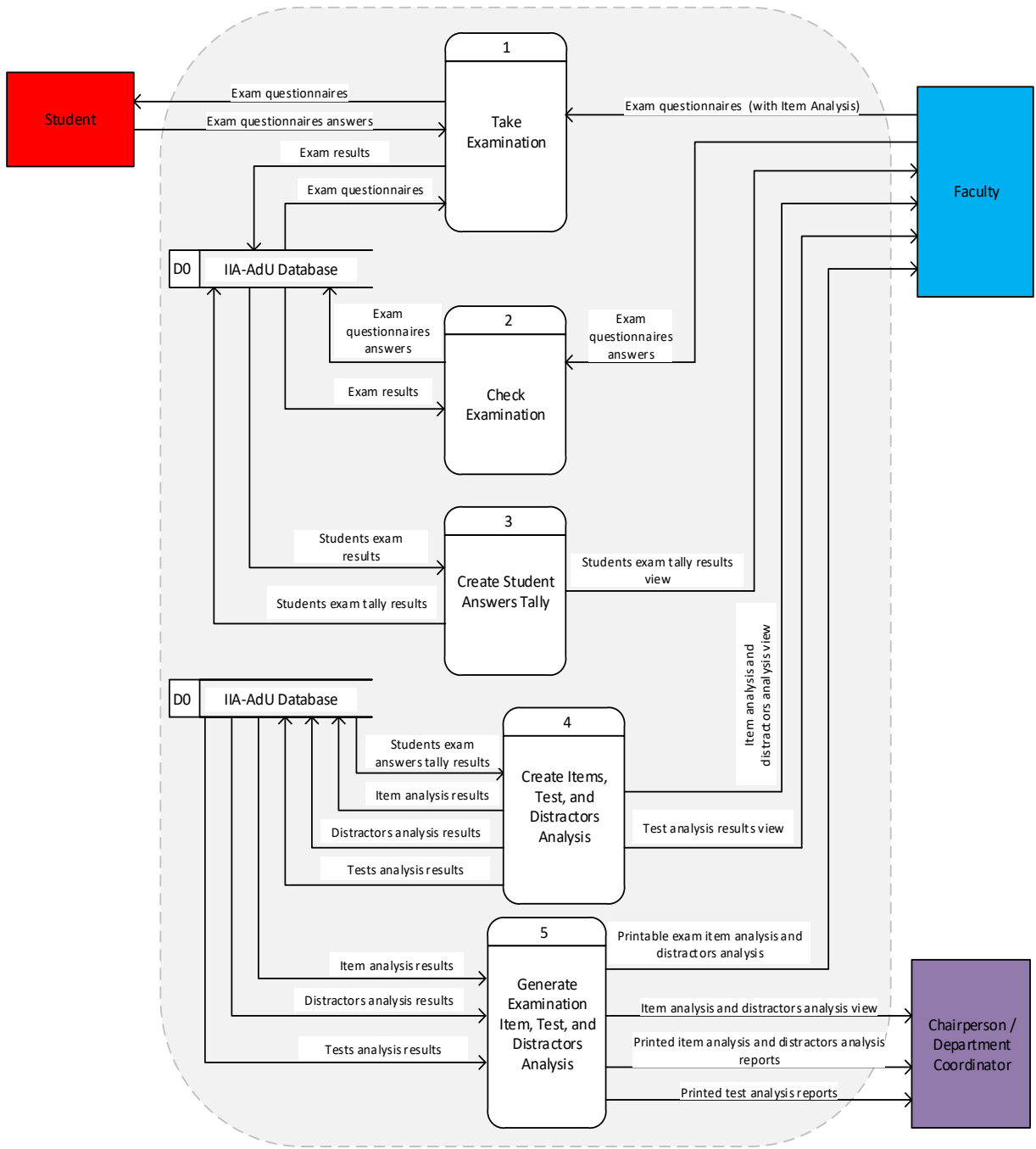

*Figure 2d.* Data Flow Diagram of Examination Management Module

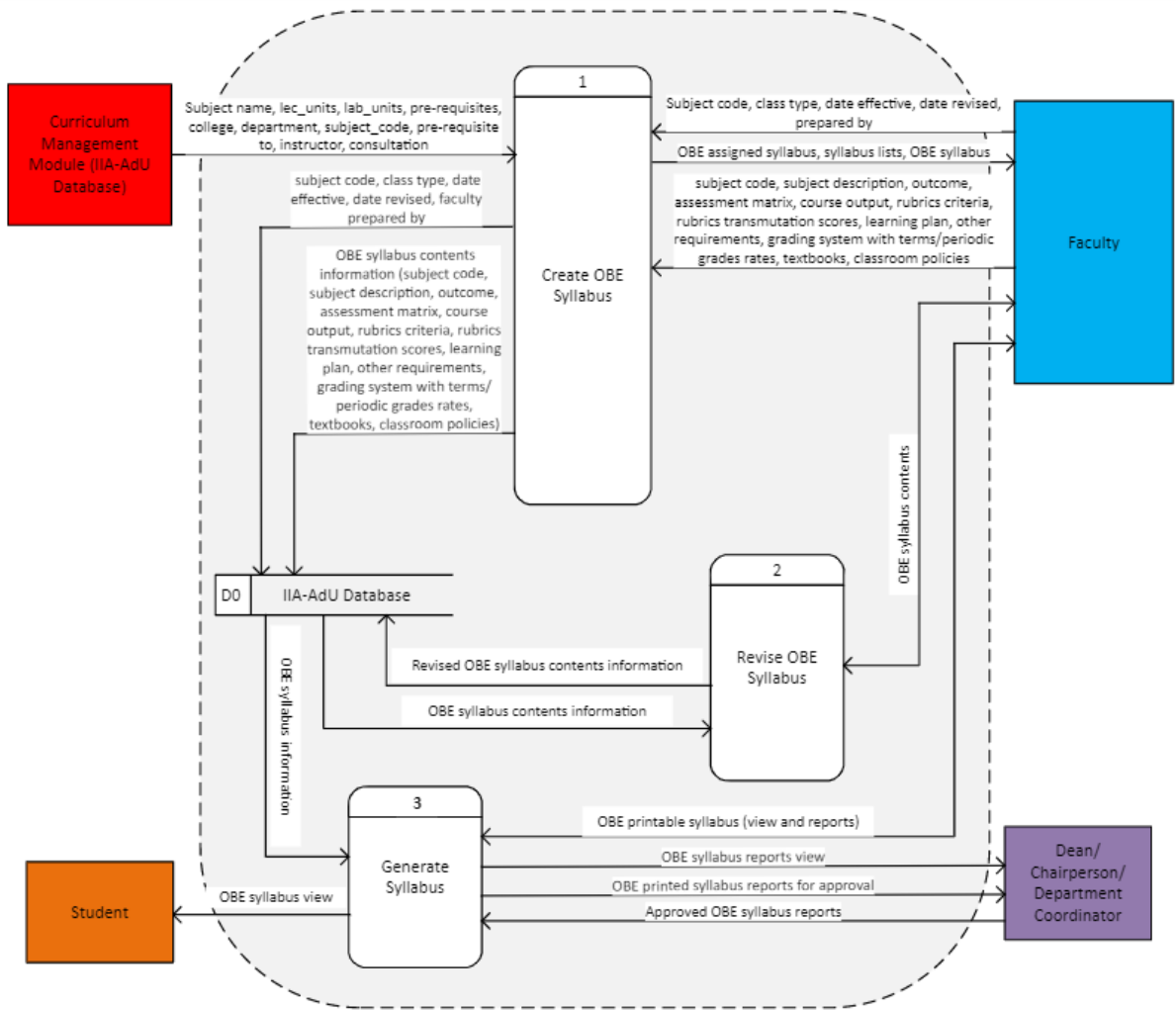

*Figure 2e.* Data Flow Diagram of Syllabus Management Module

## Data Set and Preparation

The dataset used in this study has been generated and mined from the database of the integrated academic information systems initially deployed in a specific University for one-year period and used by the selected faculty members only. The initial deployment takes place to monitor and validate the reliability, scalability, accuracy, and efficiency of the systems.

In this study, the researcher adopted a dataset extracted from the integrated academic information systems. The dataset used is intended for pure lecture and pure laboratory courses only. The main goal of data mining in this study is to come up with this dataset which used to predict the student's probability in passing their courses taken of the current semester individually based on their school performances from prelim up to midterm period.

Furthermore, the relevant data used for the prediction in terms of pure lecture and pure laboratory course is collected through attendance, class performance (quizzes, seat works, assignments, recitation, research, projects and other criteria listed in the course syllabus), prelim and midterm examination, class record id, student id, and remarks as shown in the given figure below.  Here, the data set used is composed of 82 rows (students) and 9 columns (relevant fields or attributes where the data comes from).  Out of 9 relevant attributes inside the data set, only 7 are applied in the prediction process which divided into variables.  The first one is independent variables such as prelim attendance (att_prelim), midterm attendance (att_midterm), prelim class performance (cp_prelim), midterm class performance (cp_midterm), prelim examination (exam_prelim), and midterm examination (exam_midterm), and the second one is the dependent variable (remark or passed).  The prediction variable values have been changed into binarized format where 0 is equivalent to failed, and 1 equivalent to pass the course.

The dataset used in this study were actual generated data from the assigned faculty with their chosen courses who used the system.  The weight of the data generated from the systems which set as the dataset of this study is smaller since the system was partially implemented in the university at that time being.

The dataset extracted and generated from the integrated academic systems can be downloaded in Excel format which to be uploaded to the Jupiter Notebook software for data analysis. Moreover, the dataset collected can be messy, inconsistent, and not appropriate to use in analyzing to predict students' probabilities on passing the courses. Due to this reason, data cleaning of the dataset takes place.  Data cleaning is the process of identifying and eliminating unnecessary attributes, duplicate data, data with missing values, and correcting inconsistent data.  In the first data set used in this study (pure lecture and pure laboratory data set), two attributes have been removed due to their non-importance during the data analysis processes. In addition, after the data cleaning process is done, then the dataset is ready for data analysis.

## *Proposed Data Mining Model*

Data mining holds great promise as a new and effective decision-making technique. Educational data mining, analysis, and discipline in developing methods for exploring the unique types of data from educational settings and using it for student improvement has improved.  Data mining techniques are increasingly being used in higher education to provide insights into educational and administrative problems in order to improve managerial effectiveness and the majority of educational mining research focuses on modeling student performance. Data mining is a technique that can provide teachers and students with information about each semester's academic status at the university. This technique can analyze database patterns to forecast students' performance, allowing teachers to plan a remedial program or even an actual intervention to assist and guide students who perform poorly in class.

In this study, the data mining technique used is classification. Classification is a method of categorizing objects in a dataset. It creates a model that can classify a large amount of data using a pre-classified set of data. This method consists of two processes: learning and classification. A training dataset is used in machine learning. The classification algorithm then examines the dataset. In classification, a test dataset is used to ensure that the classification rules' accuracy is acceptable.

Under this data mining technique, this study applies the decision tree algorithm to measure the prediction accuracy level. According to Jantawan and Tsai (2013) "decision tree is one of the most used techniques, due to it creates the decision tree from the data given using simple equations depending mainly on calculations of the gain ratio, which gives automatically some sort of weights to attributes/variables used and the researcher can implicitly recognize the most effective attributes on the predicted target". Furthermore, decision tree is the appropriate and practical way in building a classification model due to its relatively fast and a supervised predictive modelling approach that uses a tree-like flowchart illustration structure. To determine the best and most useful attribute in the decision tree, information gain and gini index will be utilized. Gini index is usually the measurement of impurity/purity of independent variables used in building the decision tree. Moreover, the information gain refers to the decrease in entropy after the dataset has been split as the basis of the independent variable's impurity/purity as the degree of the randomized data used.

## *Evaluation of the Model*

In this stage of the study, the researcher will evaluate the defined model in predicting the probabilities of the students to pass the courses (courses) they take in the current semester as early as possible at the time frame of prelim up to midterm period. To evaluate the defined model, there are different activities to be considered such as assessing the results towards the accurateness that the model brings towards the goal of the study, reviewing the processes which will help to determine whether the important tasks or factor of the processes has been accomplished or overlooked, and determining the next steps to undergo once the model satisfied the goal of the study. In addition, decision tree confusion matrix will be employed to measure the accuracy of the new model once it will determine.

## *Other Tools Used in the Study*

To deploy and fully implement the outcome of the study, the researcher adopted and take advantages of the open-source technology for web development, MS Excel used as dataset format extracted from the database of the systems, and Jupiter Notebook software used to analyze the dataset and obtain the measurement of accurateness of the discovered prediction model used in the study.

## RESULTS

The prediction of the student's probability to pass his current courses taken at the early stage at the time frame of prelim to midterm period used and applied a single dataset. This dataset used in this study was for pure lecture and pure laboratory courses only. The said dataset has been extracted from the database of an integrated academic information systems and exported into a Comma Separated Values (CSV) file and uploaded in a separate machine language software. This dataset was analyzed using the Jupiter Notebook software, which is phyton programming software that can be used for machine learning, data cleaning and transformation, numerical simulation, statistical modeling, and data visualization.

The researcher undergoes rigorous stages as stated below to achieve the desired goal of this study:

- The dataset was loaded in Anaconda Jupiter software.
- Has identified variables/attributes used as predictors, prediction, and unused during the data mining processes.
- Has cleaned the dataset used by deleting unused attributes and data with null values.
- Assigned the clean predicted variable to y variable which used as a storage for predicted train and test data.
- Assigned the clean predictor variables to X variable used as a storage of predictors train and test data.
- Splitting the data into X, y train and X, y test and displays the rows and columns characteristics.
- Assigned percentage of data used for X, y train and X, y test fitted in the model discovered and used in this study.
- Identified and build the model which is the Decision Tree for the intended predictive model as the goal of the study.
- Fitted the model and made the prediction of the study and results in terms of accuracy and precision, the predictive model has good predictive ability which data shown in Figure 3.

```python
#measuring the accuracy
from sklearn.metrics import confusion_matrix
from sklearn.metrics import accuracy_score
from sklearn.metrics import precision_score
from sklearn.metrics import recall_score
from sklearn.metrics import f1_score

#accuracy score
accuracy = accuracy_score(y_test, y_pred)
print('Accuracy: %f' % accuracy)

#precision score: tp / (tp + fp)
precision = precision_score(y_test, y_pred)
print('Precision: %f' % precision)

#recal score: tp / (tp + fn)
recall = recall_score(y_test, y_pred)
print('Recall: %f' % recall)

#f1 score: 2*R*P / R+P or 2 tp / (2 tp + fp + fn)
f1 = f1_score(y_test, y_pred)
print('F1 score: %f' % f1)
```

```
Accuracy: 0.761905
Precision: 0.833333
Recall: 0.882353
F1 score: 0.857143
```

*Figure 3.* Accuracy Results Measurement using the Decision Tree

- Tested the models` performance using the Confusion Matrix. As shown in Figure 4, the confusion matrix is the overall static accounting for ratio of true positive, true negative, and false negative. Thus, a "high" value essentially means that true positives and true negatives were correctly predicted.

```python
#tn  fp
#fn  tp

from sklearn.metrics import confusion_matrix
cm = confusion_matrix(y_test, y_pred)
print("Confusion Matrix : \n", cm)
```

```
Confusion Matrix :
 [[ 1  3]
 [ 2 15]]
```

*Figure 4.* Confusion Matrix Result using Decision Tree

- Identified the Decision Tree Diagram results of the prediction of the study.

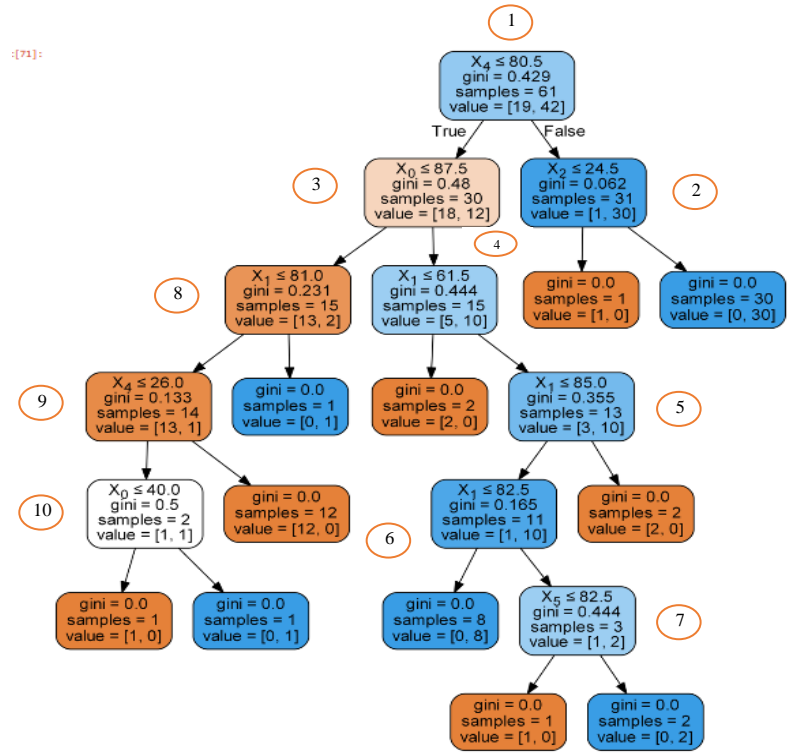

*Figure 5.* Decision Tree Results Diagram of the Prediction

To accomplish the objective of this study, CRISP-DM methodology was strictly followed. To satisfy the model's output in this study, the researcher evaluated the accurateness and precision of the prediction provided by the model used, reviewed the processed undergo to achieved the goal of the study and ensure that the model used is fitted and best model in prediction of students probabilities in passing the courses, and after being satisfied by the results of the model produced in predicting students probabilities in passing their courses currently taken at the early stage of schooling, the deployment of the study should take place and considered. Moreover, to test the accuracy of the new model, confusion matrix has been utilized. Figure 4 shows the confusion matrix results of the model determined in this study. As per confusion matrix results, the data showed that there was one student predicted to failed and the actual data showed that the student is likely failed; there were three students predicted to pass, but the actual data shows that those three students will likely to failed; that there were two students predicted to fail, but the actual data showed that those students ware passed; and that there were fifteen students predicted to passed and the actual data showed those students were passed. It is observed that a "high" value essentially means that true positives and true negatives were correctly predicted.

To predict the probabilities of the student to pass the semestral course in an early stage, Decision Tree Algorithm has been applied. To make the processes of prediction automated, Jupiter Notebook has been utilized by the researcher. Furthermore, in the given dataset, there are six independent variables (predictors) has been used and one dependent variable (prediction). The sample data inside the dataset has been split into two which likely 75% assigned for data train (data used to fit the developed model), and 25% assigned for data test (data used to validate and assess the performance of the developed model). Figure 5 show the decision tree diagram result determined by Jupiter Notebook. The decision tree diagram shows the six inputs/independent variables within the dataset. It comprised such as $X_0$ (att_prelim), $X_1$ (cp_prelim), $X_2$ (exam_prelim), $X_3$ (att_midterm), $X_4$ (cp_midterm), $X_5$ (exam_midterm) and with predicted output of pass or fail. As shown in Figure 5, a binary decision tree has been created through the process of dividing or splitting the inputs. A greedy approach is used to divide the input variables which also known as recursive binary splitting. In this approach, various split points are tried and tested using a cost function and the split with the lowest cost is chosen. The cost function is used to assess and select all input variables and split points in a greedy manner.

The GINI index which indicates how pure the nodes or split points will be utilized. In the given diagram, it illustrates the number of students predicted to passed and failed in every input variable which detects the purity of the random data used. The diagram stated as follows: (1) the samples contained in the root node is not pure, the data are not homogenous, and it is predicted that there are 19 students who will likely to fail, and 42 will pass; (2) the samples contained in node 2 are pure, both homogenous, and predicted that 1 student will fail, 30 will pass; (3) the samples contained in node 3 are not pure, not homogenous, and predicted that 13 students will fail, and 2 will pass; (4) the samples contained node 4 is predicted that 5 students will fail, 10 will pass, and out of 15 samples, 2 data are homogenous and the rest are not; (5) at node 5, it is predicted that 3 students will fail, 10 will pass and out of 13 samples, 2 data are homogenous and the rest are not; (6) at node 6, it is predicted that 1 student will fail, 10 will pass, and out of 11 sample data, 8 are homogenous and the rest is not; (7) at node 7, it is predicted that 1 student will fail, 2 will pass and the data is not pure where 1 data belong to another input, and 2 data belongs to another, and both are homogenous; (8) at node 8, it is predicted that 13 students will fail, 2 will pass and data are not pure. Out of 15 samples, 2 data are homogenous; (9) at node 9, it is predicted that 13 students will likely fail, 1 pass, and out of 14 samples, 12 data are homogenous; and (10) at node 10, it is predicted that 1 student will fail, 1 will pass, and has equal distribution of data to different inputs, but both are homogenous.

With the utilization of the model's prediction, students' chances of passing the course yields a measurement of 0.7619 accuracy, 0.8333 precision, 0.8823 recall, and 0.8571 f1 score, indicating that the determined model has good predictive ability.

## CONCLUSIONS AND RECOMMENDATIONS

Integrated Academic Information Systems has been used by many institutions which caused them to have a bulk or wide range of data stored in their databases. The discovery of new knowledge with the application of data mining by utilizing the wide range of data in the higher institutions gives great advantages to the administrators and stakeholders in terms of decision making towards students' improvement in academics. As one of the most important aspects in the aspects of the academe, the academic success of the students is a highly important issue for the management.  The early indicators towards students` probability to pass the course helps the institution administrators to take timely action to guide and improve student performance through extra coaching and counseling.  The study shows that the data mining techniques can provide effective and efficient innovative tools for analyzing and predicting student performances. In this paper, data mining is being utilized to develop a new knowledge to help learners, teacher, and education institutions administrator to monitor, guide, and uplift the quality of learning they provide to their learners in a more effective, efficient, and systematic manner.  The model used in this study will greatly affect the way educators understand and identify the weakness of their students in the class, the way they improved the effectiveness of their learning processes gearing to their students, bring down academic failure rates, and help institution administrators modify their learning system outcomes.  Furthermore, tree-based methods classify instances by sorting them down the tree from the root to a leaf node, which provides the classification of a specific instance. This paper concentrates on assessing student academic performance by leveraging data mining techniques.

To fully automate the prediction results and make them accessible to students, educators, and institution administrators for quicker management decision-making, the researcher also suggests that the prediction processes be integrated into integrated academic information systems in the future. Additionally, it is advised to incorporate automated and manual academic criteria indicators, allowing students to choose which criteria they need to meet to pass the courses they must take at the conclusion of the semester as early as the midterm period. The vast amount of data used for this kind of study will also have a significant impact on the findings of the generated predictive model, which have the highest accuracy rate.

## IMPLICATIONS

The discovery of new knowledge with the application of data mining through utilization of wide range of data in the higher institutions gives great advantages to the administrators and stakeholders in terms of decision making towards students' improvement in academics. With the implementation of an integrated academic information systems with predictive model for student`s probability of passing semestral course, early indicators towards students` probability to pass the subjects are automatically defined which helps the institution administrators to take timely action to guide and improve student performance through extra coaching and counseling.

## ACKNOWLEDGEMENT

Deepest gratitude to Dr. Arnel M. Avelino, Dean - College of Engineering, Computer Studies, and Architecture, to the Research and Knowledge Management Department, and to the administration of LPU-Cavite for their outmost support and inspiration to their researchers.

## DECLARATIONS

### *Conflict of Interest*

The author declares that there is no conflict of interest.

### *Informed Consent*

The data used in this study was collected from the participants with consent and well informed to its purpose. The participant's identity was not revealed and not included in the dataset used in this study.

### *Ethics Approval*

The LPU-Cavite Center for Research and Knowledge Management and the Research Ethics Committee accepted and approved the conduct of the study.

**Author's Biography**

Anabella C. Doctor is a Computer Engineering Department chair at Lyceum of the Philippines University – Cavite. She finished several research focused on software development and related fields, presented in both local and international conferences with journal publication which was funded by a university and Commission on Higher Education. She finished master's in Information Technology at TUP, Manila and currently taking Doctor of Information Technology at DLSU - Dasmariñas. She is proficient web programming, Visual C# and Visual Basic programming language with MS SQL for database. She is also an author of two computer books used in junior high school.